\documentclass[10pt,twocolumn,letterpaper]{article}

\usepackage{cvpr}              

\usepackage{graphicx}
\usepackage{amsmath}
\usepackage{amssymb}
\usepackage{booktabs}
\usepackage{multirow}
\usepackage{stfloats}
\usepackage{bbding}
\usepackage{color}
\usepackage{threeparttable}
\usepackage{graphicx}
\usepackage{wrapfig}
 \usepackage{mathtools}

\usepackage{appendix}

\usepackage{axessibility}

\usepackage[marginal]{footmisc}

\usepackage{color}
\usepackage{hyperref}
\hypersetup{
	colorlinks=true,
	linkcolor=red,
	filecolor=blue,      
	urlcolor=red,
	citecolor=green,
}

\usepackage[capitalize]{cleveref}
\crefname{section}{Sec.}{Secs.}
\Crefname{section}{Section}{Sections}
\Crefname{table}{Table}{Tables}
\crefname{table}{Tab.}{Tabs.}

\def\cvprPaperID{6719} 
\def\confName{CVPR}
\def\confYear{2022}

\begin{document}

\title{Learning to Answer Questions in Dynamic Audio-Visual Scenarios}
\vspace{-10mm}
\author{
\textbf{Guangyao Li}\textsuperscript{1,$\dagger$}, 
\textbf{Yake Wei}\textsuperscript{1,$\dagger$}, 
\textbf{Yapeng Tian}\textsuperscript{3,$\dagger$}, 
\textbf{Chenliang Xu}\textsuperscript{3}, 
\textbf{Ji-Rong Wen}\textsuperscript{1}, 
\textbf{Di Hu}\textsuperscript{1,2,*}
\vspace{0.5em}
\\
\textsuperscript{1}Gaoling School of Artificial Intelligence, Renmin University of China, Beijing\\
\textsuperscript{2}Beijing Key Laboratory of Big Data Management and Analysis Methods, Beijing\\
\textsuperscript{3}Department of Computer Science, University of Rochester, Rochester
\\
\textsuperscript{1}\{guangyaoli, yakewei, jrwen, dihu\}@ruc.edu.cn, \textsuperscript{3}\{yapengtian, chenliang.xu\}@rochester.edu
}

\maketitle

\vspace{-1em}
\begin{abstract}
In this paper, we focus on the Audio-Visual Question Answering (AVQA) task, which aims to answer questions regarding different visual objects, sounds, and their associations in videos. The problem requires comprehensive multimodal understanding and spatio-temporal reasoning over audio-visual scenes. To benchmark this task and facilitate our study, we introduce a large-scale MUSIC-AVQA dataset, which contains more than 45K question-answer pairs covering 33 different question templates spanning over different modalities and question types. We develop several baselines and introduce a spatio-temporal grounded audio-visual network for the AVQA problem. Our results demonstrate that AVQA benefits from multisensory perception and our model outperforms recent A-, V-, and AVQA approaches. We believe that our built dataset has the potential to serve as testbed for evaluating and promoting progress in audio-visual scene understanding and spatio-temporal reasoning. 
Code and dataset: 
\href{http://gewu-lab.github.io/MUSIC-AVQA/}
{http://gewu-lab.github.io/MUSIC-AVQA/}
\footnote{
\textsuperscript{$\dagger$}Equal contribution. 
\textsuperscript{*}Corresponding author.
}
\end{abstract}

\vspace{-1.5em}
\section{Introduction}
\label{sec:intro}
\vspace{-0.5em}
We are surrounded by audio and visual messages in daily life, and both modalities jointly improve our ability in scene perception and understanding~\cite{holmes2005multisensory}.
For instance, imagine that we are in a concert, watching the performance and listening to the music at the same time contribute to better enjoyment of the show.
Inspired by this, how to make machines integrate multimodal information, especially the natural modality such as the audio and visual ones, to achieve considerable scene perception and understanding ability as humans is an interesting and valuable topic.

\begin{figure}[t]
     \centering
     \includegraphics[width=0.45\textwidth]{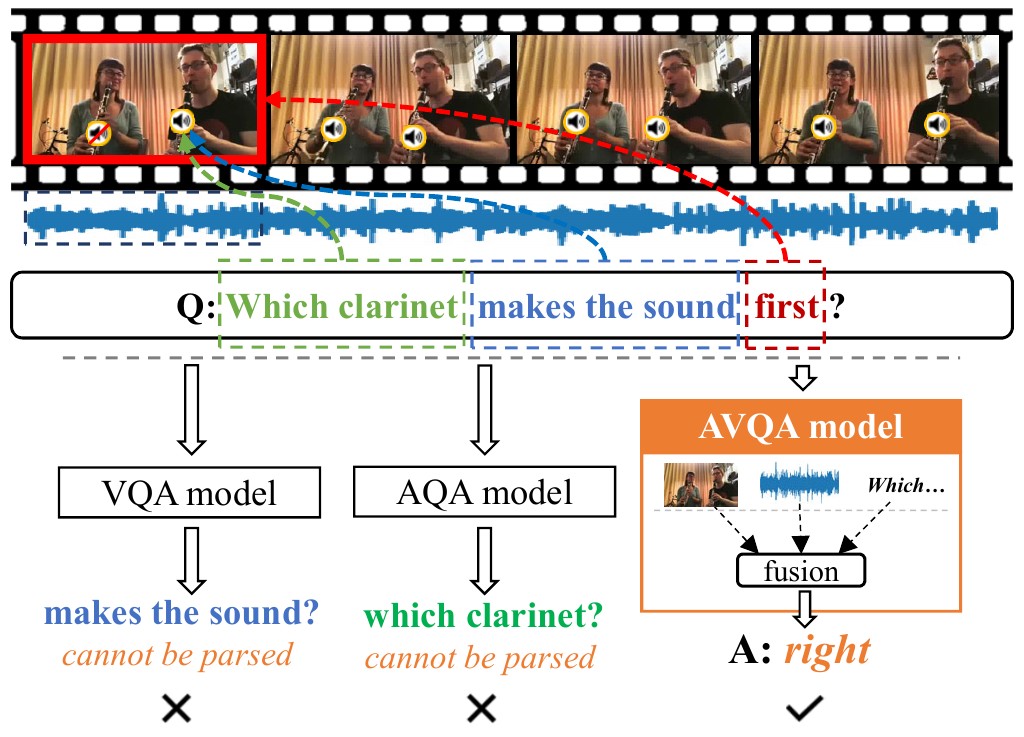}
     \vspace{-1.9mm}
     \caption{Audio-visual question answering requires auditory and visual modalities for multimodal scene understanding and spatio-temporal reasoning. For example, when we encounter a complex musical performance scene involving multiple sounding and non-sounding instruments above, it is difficult to analyze the \textit{sound first} term in the question by VQA model that only considers visual modality. While if we only consider the AQA model with mono sound, the \textit{left} or \textit{right} position is also hard to be recognized. However, we can see that using both auditory and visual modalities can answer this question effortlessly.}
     \label{fig:teaser}
     \vspace{-5.8mm}
\end{figure}

In recent years, we have seen significant progress in sounding object perception~\cite{senocak2018learning, hu2020discriminative,Chen_2021_CVPR,Yang_2021_ICCV}, audio scene analysis~\cite{zhao2018sound, Gao_2021_CVPR, xu2019recursive,  hu2019deep, gan2020music, hu2020cross, diheterogeneous}, audio-visual scene parsing~\cite{tian2020unified, wu2021exploring}, and content description~\cite{xu2017learning,tian2019audio, iashin2020multi} towards audio-visual scene understanding. Although these methods associate objects or sound events across audio and visual views, most of them remain limited ability for cross-modal reasoning, under complex audio-visual scenarios. In contrast, humans are capable of performing multi-step spatial and temporal reasoning over multimodal contexts to solve complex tasks, such as answering an audio-visual question, but it is quite challenging for machines. Existing methods such as \emph{Visual Question Answering} (VQA)~\cite{antol2015vqa} and \emph{Audio Question Answering} (AQA)~\cite{fayek2020temporal} only focus on single modality, which cannot reason well in a more natural scenario with both audio and visual modalities. For instance, as shown in Fig.~\ref{fig:teaser}, when answering the audio-visual question ``\textit{Which clarinet makes the sound first}" for this instrumental ensemble, it requires to locate sounding objects ``\textit{clarinet}" in the audio-visual scenario and focus on the ``\textit{first}" sounding ``\textit{clarinet}" in the timeline. To answer the question correctly, both effective audio-visual scene understanding and spatio-temporal reasoning are essentially desired.

In this work, we focus on the \emph{Audio-Visual Question Answering} (AVQA) task, which aims to answer questions regarding visual objects, sounds and their association. To this end, a computational model is essentially required to equip with effective multimodal understanding and reasoning ability on rich dynamic audio-visual scenes. To facilitate the aforementioned research, we built a large-scale \emph{Spatio-Temporal Music} AVQA (MUSIC-AVQA) dataset.
Considering that musical performance is a typical multimodal scene consisting of abundant audio and visual components as well as their interaction, it is appropriate to be utilized for the exploration of effective audio-visual scene understanding and reasoning. So we collected amounts of user-uploaded videos of musical performance from YouTube, and videos in the built dataset consist of solo, ensemble of the same instruments and ensemble of different instruments. It contains 9,288 videos covering 22 instruments, with a total duration of over 150 hours. 45,867 question-answer pairs are generated by human crowd-sourcing, with an average of about 5 QA pairs per video. The questions are derived from 33 templates and asked regarding content from different modalities at space and time, which are suitable to explore fine-grained scene understanding and spatio-temporal reasoning in the audio-visual context.

To solve the above AVQA task, we consider this problem from the spatial and temporal grounding perspective, respectively. Firstly, the sound and the location of its visual source is deemed to reflect the spatial association between audio and visual modality, which could help to decompose the complex scenario into concrete audio-visual association. Hence, we propose a spatial grounding module to model such cross-modal association through attention-based sound source localization. Secondly, since the audio-visual scene changes over time dynamically, it is critical to capture and highlight the key timestamps that are closely related to the question. Accordingly, the temporal grounding module that uses question features as queries is proposed to attend crucial temporal segments for encoding question-aware audio and visual embeddings effectively. Finally, the above spatial-aware and temporal-aware audio-visual features are fused to obtain a joint representation for Question Answering. As an open-ended problem, the correct answers to questions can be predicted by choosing words from a pre-defined answer vocabulary. Our results indicate that audio-visual QA benefits from effective audio-visual scene understanding and spatio-temporal reasoning, and our model outperforms recent A-, V-, and AVQA approaches. 

To summarize, our contributions are threefold:

\begin{itemize}   
\vspace{-0.85em}
\item We build the large-scale MUSIC-AVQA dataset of musical performance, which contains more than 9K videos annotated by over 45K QA pairs, spanning over different modal scenes.
\vspace{-0.85em}
\item A spatio-temporal grounding model is proposed to solve the fine-grained scene understanding and reasoning over audio and visual modalities.
\vspace{-0.85em}
\item Extensive experiments show that AVQA benefits from multisensory perception and our model is superior to recent QA approaches especially on the questions that measures spatio-temporal reasoning ability of models.
\end{itemize}

\vspace{-0.6em}
\section{Related Work}
\label{sec:relatedwork}
\vspace{-0.3em}
\subsection{Audio-Visual Learning}
\vspace{-0.5em}
By integrating the audio and visual information in multimodal scenes, it is expected to explore more sufficient scene information and overcome the limited perception in single modality. Recently, there have been several works utilizing audio and visual modality to facilitate multimodal scene understanding in different perspectives, such as sound source localization~\cite{senocak2018learning, wu2019dual, qian2020multiple,liu2022visual, hu2021class} and separation~\cite{zhao2018sound, Gao_2021_CVPR, gan2020music,tian2021cyclic,zhou2022sepfusion, zhou2020sep}, audio inpainting~\cite{zhou2019vision}, event localization~\cite{tian2018audio, brousmiche2021multi, zhou2021positive}, action recognition~\cite{gao2020listen}, video parsing~\cite{tian2020unified, wu2021exploring}, captioning~\cite{xu2017learning, tian2019audio, iashin2020multi}, and dialog~\cite{alamri2019audio, zhu2020describing}. 

Regarding previous works on sound source localization and separation, the former mainly focuses on locating sounds in a visual context~\cite{senocak2018learning, qian2020multiple}, while the latter mainly centers around separating different sounds from corresponding visual objects~\cite{zhao2018sound, gao2018learning}.
These works have made great progress for the interaction of audio and visual features, but they essentially focus on the perception of audio-visual objects.
Further, some researchers propose to integrate audio and visual messages to explore semantic events and behaviors in multimodal scenes~\cite{tian2018audio, gao2020listen}. As expected, these works have shown considerable performance by utilizing more sufficient information from audio and visual cues.
Based on which, others took a step forward to parse the audio-visual scenes~\cite{tian2020unified},
describe content~\cite{iashin2020multi}, and leverage contextual cues for dialog~\cite{alamri2019audio, zhu2020describing}. 

Apart from the above methods that facilitate scene understanding by excavating and analyzing different modalities, a unified multimodal model should also be able to reason their spatio-temporal correlation. In this work, different from the previous methods, besides the fine-grained scene understanding, we further propose to explore spatio-temporal reasoning in the audio-visual context.

\begin{table*}[h]
\begin{center}

\scalebox{0.66}{
\begin{tabular}{c|c|c|c|c|c|c|ccccc}
\hline
\multirow{2}{*}{Dataset} & \multirow{2}{*}{Origin} & \multirow{2}{*}{Main sound type} & \multirow{2}{*}{\# Videos} & \multirow{2}{*}{\begin{tabular}[c]{@{}c@{}}Average\\ video length\end{tabular}} & \multirow{2}{*}{A Question} & \multirow{2}{*}{V Question} & \multicolumn{5}{c}{A-V Question}                                                                                                                    \\
                         &                         &                          &                            &                                    &                             &                             & Existential                 & Location                    & Counting                    & Comparative                 & Temporal                    \\ \hline
ActivityNet-QA~\cite{yu2019activitynet}           & ActivityNet             & Background music         & 5.8K                       & 180s                                & \XSolidBrush & \Checkmark   & \XSolidBrush & \XSolidBrush & \XSolidBrush & \XSolidBrush & \XSolidBrush \\
TVQA~\cite{lei2018tvqa}                     & TV Show                 & Human speech             & 21.8K                      & 60s/90s                              & \XSolidBrush & \Checkmark   & \XSolidBrush & \XSolidBrush & \XSolidBrush & \XSolidBrush & \XSolidBrush \\
AVSD~\cite{alamri2019audio}                     & Charades                & Domestic sounds          & 8.5K                       & 30s                                 & \Checkmark & \Checkmark   & \Checkmark   & \XSolidBrush & \XSolidBrush & \XSolidBrush & \XSolidBrush \\
Pano-AVQA~\cite{yun2021pano}                & Online                  & Visual object sound      & 5.4k                       & 5s                                  & \Checkmark & \Checkmark   & \Checkmark   & \Checkmark   & \XSolidBrush & \XSolidBrush & \XSolidBrush \\ \hline
MUSIC-AVQA                & YouTube                 & Visual object sound      & 9.3K                       & 60s                                 & \Checkmark   & \Checkmark   & \Checkmark   & \Checkmark   & \Checkmark   & \Checkmark   & \Checkmark   \\ \hline
\end{tabular}
}
\vspace{-0.75em}
\caption{\textbf{Comparison with other video QA datasets.} Our MUSIC-AVQA dataset focuses on the interaction between visual objects and their produced sounds, offering QA pairs that cover audio, visual and audio-visual questions, which is more comprehensive than other datasets. The collected videos in MUSIC-AVQA can facilitate audio-visual understanding in terms of spatial and temporal associations. 
}
\vspace{-9mm}
\label{dataset}


\end{center}
\end{table*}

\vspace{-0.25em}
\subsection{Question Answering}
\vspace{-0.5em}
In the past years, several question answering tasks have been proposed but in different modalities, including text question answering~\cite{weston2015towards, rajpurkar2016squad}, visual question answering~\cite{antol2015vqa, jang2017tgif, yu2015visual, zhang2016yin}, audio question answering~\cite{zhang2017speech, fayek2020temporal}, etc.

VQA ~\cite{antol2015vqa, lu2016hierarchical, goyal2017making} aims to generate natural language answers about specific visual content. 
The early research in VQA focused on simple visual understanding in static images but ignored the spatial and semantic relationships between visual content, hence they are difficult to achieve effective visual reasoning in complex scene.
To overcome this shortcoming, Johnson~\emph{et al.}~\cite{johnson2017clevr} released the simulated CLEVR dataset and expected the model to answer reasoning-oriented visual questions.
Since then, more attentions are paid to the spatial and semantic relational reasoning of visual objects in VQA ~\cite{gao2019dynamic, norcliffe2018learning, andreas2016neural}.
Recently, some methods proposed to improve the spatial-temporal reasoning ability of computational model further, by answering question in the video context~\cite{zhao2017video, kim2020modality, fan2019heterogeneous, yu2019activitynet, li2019beyond, Xiao_2021_CVPR}.
Apart from the visual information, some other modality information in video, such as subtitles~\cite{lei2018tvqa} or scripts~\cite{tapaswi2016movieqa}, are used for advancing the understanding of video content. Similarly, some external knowledge~\cite{wu2017image, garcia2020knowledge} and situations~\cite{castro2020lifeqa, wu2021star} are also utilized to achieve better content understanding.

In addition to the visual modality-based QA, some researchers also proposed to answer questions in other modalities, such as audio~\cite{fayek2020temporal, schwartz2019simple, alamri2019audio,yun2021pano} and speech~\cite{zhang2017speech}.
Pano-AVQA~\cite{yun2021pano} is a concurrent work to ours, also aiming at audio-visual question answering. 
But the QA-pairs within the dataset only covers relatively simple audio-visual association, such as \textit{existential} or \textit{location} questions. 
In contrast, our built MUSIC-AVQA dataset can facilitate study on spatio-temporal reasoning for dynamic and long-term audio-visual scenes. Meanwhile, the proposed method provides new perspectives in modeling such complex scenario and obtains noticeable results.

\vspace{-0.3em}
\section{The MUSIC-AVQA Dataset}
\label{sec:dataset}
\vspace{-0.25em}
\subsection{Overview}
\vspace{-0.5em}
To explore scene understanding and spatio-temporal reasoning over audio and visual modalities, we build a large-scale audio-visual dataset, MUSIC-AVQA, which focuses on question-answering task. As noted above, high-quality datasets are of considerable value for AVQA research. Hence, considering that musical performance is a typical multimodal scene consisting of abundant audio and visual components as well as their interaction, we choose to manually collect amounts of musical performance videos from YouTube. Specifically, 22 kinds of instruments, such as guitar, cello, and xylophone, are selected and 9 audio-visual question types are accordingly designed, which cover three different scenarios, \emph{i.e.}, audio, visual and audio-visual. 

As shown in Tab.~\ref{dataset}, compared to existing related datasets, our released MUSIC-AVQA dataset has the following advantages:
\textbf{1)} Our dataset offers QA pairs that covering audio question, visual question and audio-visual question, which is more comprehensive than other datasets. Most video QA datasets, like ActivityNet-QA~\cite{yu2019activitynet}, TVQA~\cite{lei2018tvqa}, only contain visual question and provide limited possibility to explore audio-visual correlation. Although existing AVQA datasets, such as AVSD~\cite{alamri2019audio} and Pano-AVQA~\cite{yun2021pano}, also offer audio-visual QA pairs, they focus on relatively simple audio-visual correlation that only needs spatial reasoning, such as \textit{existential} or \textit{location} questions. As a concurrent work of Pano-AVQA, our dataset is more comprehensive and much longer than it, which includes more spatial and temporal related question, such as \textit{existential}, \textit{location}, \textit{counting}, \textit{comparative} and \textit{temporal}. 
\textbf{2)} Our dataset consists of musical performance scenes that contains enriching audio-visual components, which contributes to better investigation of audio-visual interaction, and it can avoid the noise problem in the scene to some extent, where the visual objects and sounds are not related. 
The audio information in most released datasets (\eg, ActivityNet-QA~\cite{yu2019activitynet} and AVSD~\cite{alamri2019audio}) is usually accompanied by severe noise that sound and visual objects in the video do not match (\eg. background music), which makes them difficult to explore the association between different modalities. In addition, the TVQA~\cite{lei2018tvqa} dataset contains both visual and audio modality, but its sound mainly consists of human speech, and only the corresponding subtitle is used during QA pairs construction. In the followings, we provide detailed descriptions about the procedure of video collection, QA pairs annotation and collection, as well as the related statistical analysis about our MUSIC-AVQA dataset.

\begin{figure*}[t]
     \centering
     \includegraphics[width=0.98\textwidth]{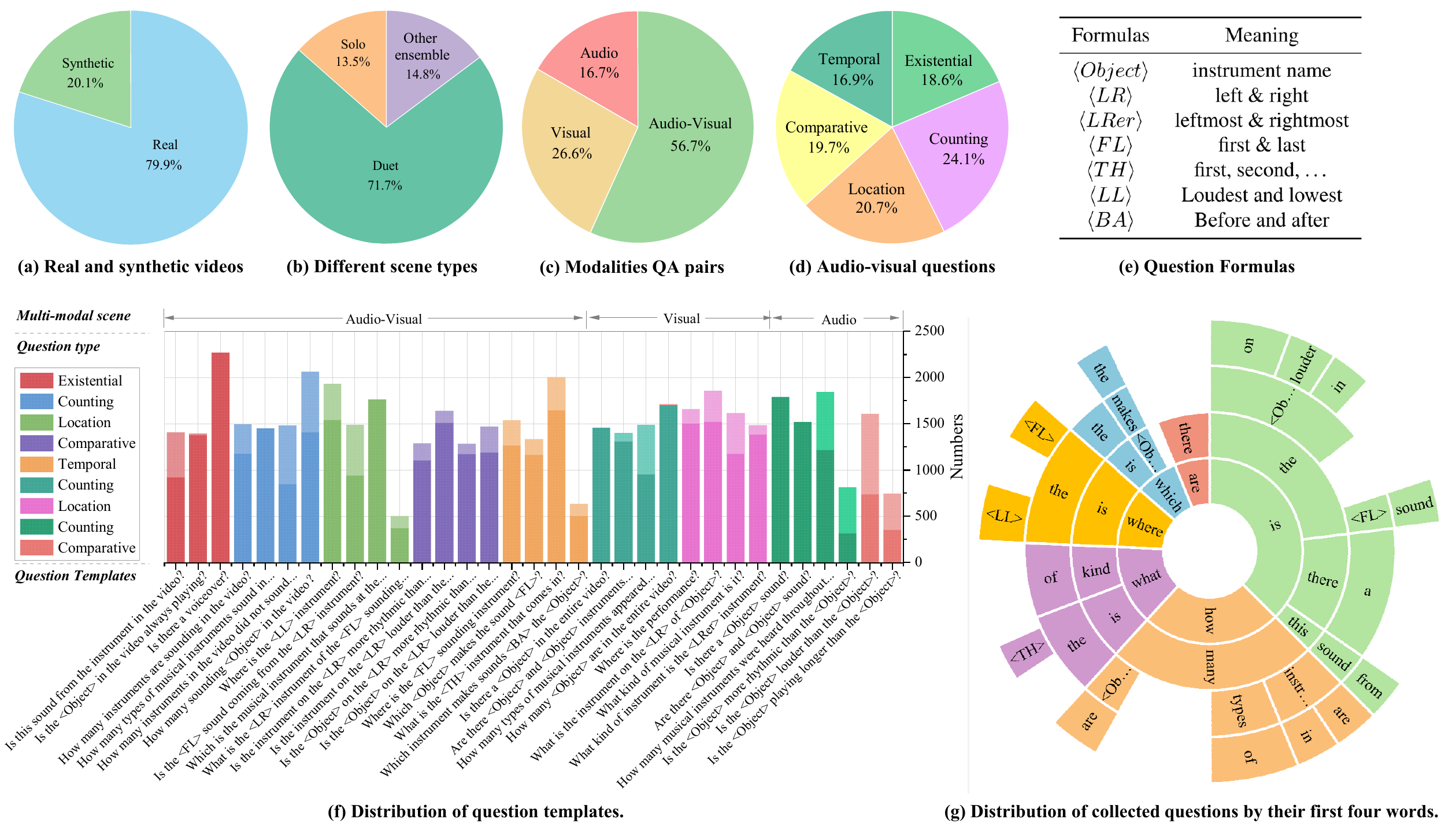}
     \vspace{-1em}
     \caption{\textbf{Illustrations of our MUSIC-AVQA dataset statistics.} \textbf{(a-d)} statistical analysis of the videos and QA pairs. \textbf{(e)} Question formulas. \textbf{(f)} Distribution of question templates, where the dark color indicates the number of QA pairs generated from real videos while the light-colored area on the upper part of each bar means that from synthetic videos. \textbf{(g)} Distribution of first n-grams in questions. Our QA-pairs need fine-grained scene understanding and spatio-temporal reasoning over audio and visual modalities to be solved. For example, \textit{existential} and \textit{location} questions require spatial reasoning, and \textit{temporal} questions require temporal reasoning. Best viewed in color.}
     \label{fig:2}
     \vspace{-5.5mm}
\end{figure*}

\subsection{Video Collection}
\vspace{-0.5em}
\noindent
\textbf{Real Videos.}
We collect 7,422 real videos of musical performance from YouTube.
Among these videos, three kinds of musical performance are covered to ensure the diversity, complexity and dynamic of audio-visual scenes: solo, ensemble of the same instrument (ESIT) and ensemble of different instruments (EDIT).
In order to control the quantity balance of different instrument types, we design the following rules:
\textbf{1) Solo}: about 50 solo videos are collected per instrument;
\textbf{2) ESIT}: about 100 videos are collected per ESIT type;
\textbf{3) EDIT}: each instrument is required to combine with every other instruments.
For the collected untrimmed videos, we randomly cut them into one minute long for efficiency purpose.
Moreover, human verification is performed to ensure whether the cut videos contain musical performance scenes.

\noindent
\textbf{Synthetic Videos.} 
There are many solo and duet performance in real-world videos that contain limited visual objects and sounds. 
To further facilitate study on understanding and reasoning,
we synthesize more challenging videos in which multiple visual objects and sounds are appeared with different associations.

\vspace{-0.15em}
\subsection{QA Pairs Annotation and Collection}
\vspace{-0.5em}
For the collected musical performance videos, the QA annotation is performed in three steps: question design, question collection and answer collection.

\vspace{-0.05em}
\noindent
\textbf{Questions Design}. 
In order to better explore the contribution of the spatio-temporal correlation between visual and audio components to multimodal scene understanding, 33 question templates that cover 9 question types are proposed under different modality scenes.
Concretely, to prevent from asking multiple simple questions and guarantee the diversity of questions, inspired by the mechanism of question templates in building VQA dataset~\cite{johnson2017clevr, talmor2021multimodalqa}, we design several question templates before annotating the collected videos, as shown in Fig.~\ref{fig:2}(d). 

\noindent
\textbf{Questions Collection}. 
We design an audio-visual question answering labeling system to collect questions.
To ensure the diversity and balance of different question templates, we set up the following rules for the labeling system:
1) the same question template in a video can only be annotated by the same annotator once;
2) each video needs to be watched for more than 30-seconds before it can be annotated;
3) the question templates that have been annotated will no longer be displayed to the subsequent annotators;
4) each video has to be annotated for 5 times.
With these rules, we collect the questions for all the musical performance videos.

\noindent
\textbf{Answers.}
As each question template has certain answer, we ask annotators to directly choose the correct one from the answer vocabulary. And we also use the above labeling system to collect answers.
In this process, we set up the following rules when answering questions:
1) when one answer that is selected for the same question twice, it will be considered as the correct answer;
2) when the answer to a question is confirmed, it will not be seen by the subsequent annotators.
In addition, the unreasonable question is annotated as invalid, and the corresponding video will be asked one new question again.

\subsection{Statistical Analysis}
\vspace{-0.5em}
Our MUSIC-AVQA dataset contains 45,867 question-answer pairs, distributed in 9,288 videos for over 150 hours. Figure~\ref{fig:2}(a-d) provides the statistical analysis of our dataset. In this dataset, real videos and synthetic videos accounted for 79.9\% and 20.1\%, respectively. Real videos are composed of 14.8\% solo videos, 71.7\% duet videos and 13.5\% other ensemble videos. Audio-visual questions makes up the majority of all QA pairs and consists of five types with a balanced share. Fig.~\ref{fig:2}(f) shows that all QA pairs types are divided into 3 modal scenarios, which contain 9 question types and 33 question templates. Finally, as an open-ended problem of our AVQA tasks, all 42 kinds of answers constitute a set for selection. For training and evaluation, we randomly split the dataset into training, validation, and testing sets with 32,087, 4,595, and 9,185 QA pairs, respectively. More details about the dataset construction and statistical analysis are in the \emph{Supp. Materials}.

\section{Method}
\begin{figure*}[t]
     \centering
     \includegraphics[width=0.97\textwidth]{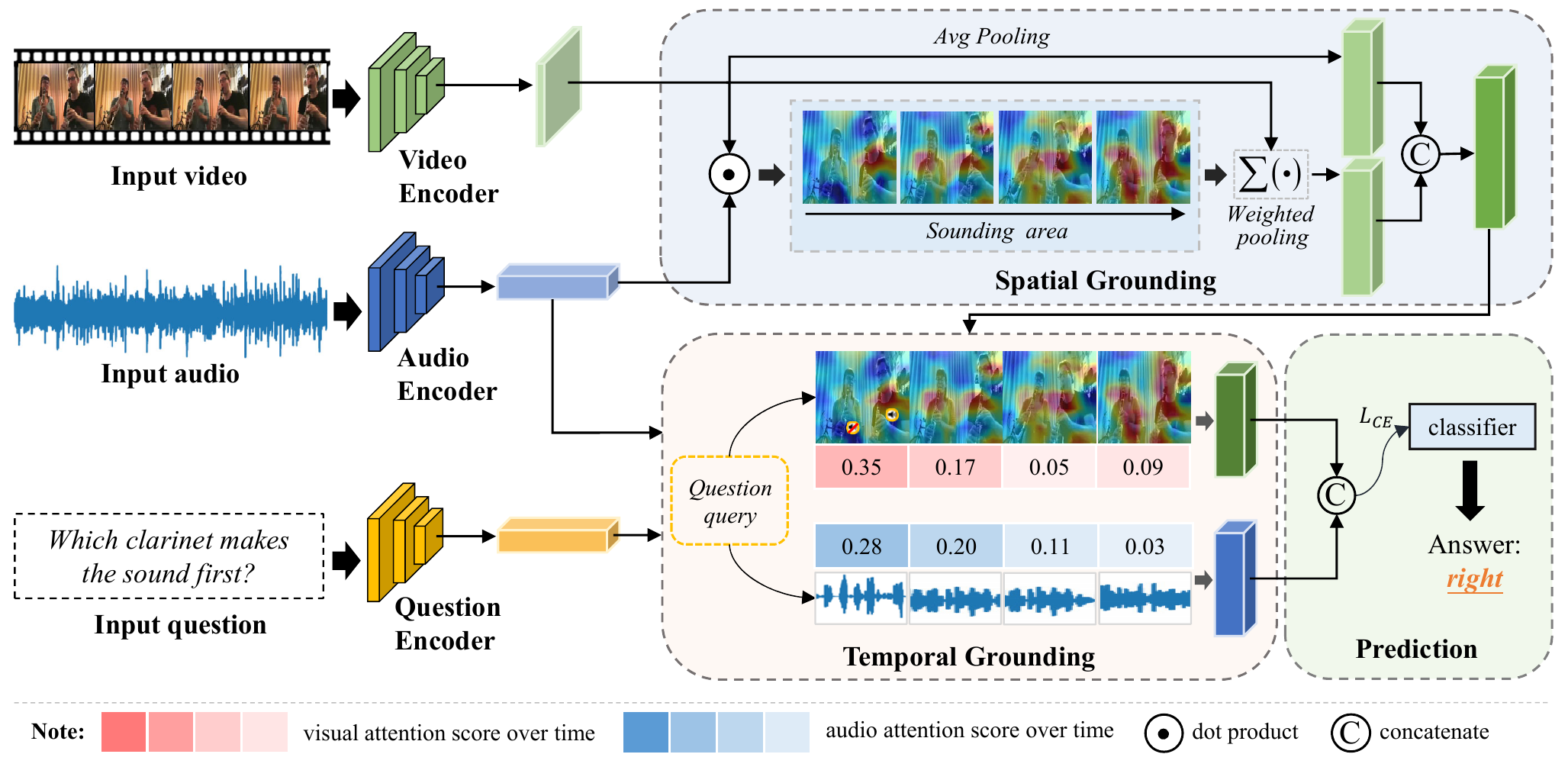}
     \vspace{-1em}
     \caption{\textbf{The proposed audio-visual question answering model.} The model takes pre-trained CNNs to extract audio and visual features and uses a LSTM to obtain a question embedding. We associate specific visual locations with the input sounds to perform spatial grounding, based on which audio and visual features of key timestamps are further highlighted via question query for temporal grounding. Finally, multimodal fusion is exploited to integrate audio, visual, and question information for predicting the answer to the input question.}
     \label{fig:framework}
     \vspace{-1.2em}
\end{figure*}

\vspace{-0.2em}
\label{sec:mehtod}
To solve the AVQA problem, we propose a spatio-temporal grounding model to achieve scene understanding and reasoning over audio and visual modalities. An overview of the proposed framework is illustrated in Fig.~\ref{fig:framework}.

\subsection{Representations for Different Modalities}
\vspace{-0.4em}
Given an input video sequence containing both visual and audio tracks, we first divide it into $T$ non-overlapping visual and audio segment pairs $\{V_t, A_t\}_{t=1}^{T}$, where each segment is $1s$ long. The question sentence $Q$ is tokenized into $N$ individual words $\{q_n\}_{n=1}^{N}$.

\noindent
\textbf{Audio Representation.} We encode each audio segment $A_t$ into a feature vector $f_a^t$ using a pre-trained VGGish model~\cite{gemmeke2017audio}, which is VGG-like 2D CNN network, employing over transformed audio spectrograms. The audio representation is extracted offline and the model is not fine-tuned.

\noindent
\textbf{Visual Representation.} We sample a fixed number of frames for all video segments. We then apply pre-trained ResNet-18~\cite{he2016deep} on video frames to extract visual feature map $f_{v,m}^t$ for each video segment $V_t$. The used pre-trained ResNet-18 model is not fine-tuned.

\noindent
\textbf{Question Representation.} For an asked question $Q = \{q_n\}_{n=1}^{N}$, a LSTM is used to process projected word embeddings $\{f_q\}_{n=1}^{N}$ and encode the question into a feature vector $f_q$ using the last hidden state. The question encoder is trained from the scratch. 

\vspace{-0.4em}
\subsection{Spatial Grounding Module}
\vspace{-0.5em}
We consider that the sound and the location of its visual source usually reflects the spatial association between audio and visual modality, the spatial grounding module, which performs attention-based sound source localization, is therefore introduced to decompose the complex scenarios into concrete audio-visual association. Specifically, for each video segment $V_t$, the visual feature map $f_{v,m}^t$ and the corresponding audio feature $f_a^t \in \mathcal{R}^{C} $ compose the matched pair. Then we randomly sample another visual segment and get its visual feature map, which composes the non-matched pair with the audio feature $f_a^t$ . For each pair, we can compute the sound-related visual features, $f_{v,s}^t$, as: 
\vspace{-2em}
\begin{align}
\label{attention}
    \vspace{-0.4em}
    f_{v,s}^t= f_{v,m}^t \cdot {\sigma}((f_a^t) ^{\intercal}\cdot f_{v,m}^t ),
    \vspace{-0.4em}
\end{align}
where $\sigma$ is the softmax and $(\cdot)^{\intercal}$ represents the transpose operator. To prevent possible visual information loss, we averagely pool the visual feature map $f_{v,m}^t$, obtaining the global visual feature $f_{v,g}^t$. The two visual feature is fused as the visual representation: $f_{v}^t = \text{\textbf{FC}} (\text{Tanh}[f_{v,g}^t,f_{v,s}^t])$,
where \textbf{FC} represents fully-connected layers.
Then, the visual and the audio representation combines to predict the audio-visual pairs are matched or not:
\vspace{-0.3em}
\begin{align}
\label{visual_fuse}
    \hat{y}^{t}= \sigma(\text{\textbf{FC}}(\text{Concat}[f_a^t, f_{v}^t])),
\end{align}
\vspace{-1.8em}
\begin{align}
\label{loss}
\mathcal{L}_{s} = \mathcal{L}_{ce}(y^{match},\hat{y}^t),
\end{align}
where $y^{match}$ indicates whether the audio and visual feature come from the matched pair, i.e., $y^{match}=1$ when $f_{v}^t$ and $f_a^t$ is the matched pair, otherwise $y^{match}=0$. $\mathcal{L}_{ce}$ is the cross-entropy loss. It should be noted that non-matched pairs are only used in the spatial grounding module, i.e., $f_{v}^t$ and $f_a^t$ is always the matched pair in other modules.

\subsection{Temporal Grounding Module} 
\vspace{-0.3em}
\label{sec:qgta}
To highlight the key timestamps that are closely associated to the question, we propose a temporal grounding module, which is designed for attending critical temporal segments among the changing audio-visual scenes and capturing question-aware audio and visual embeddings. 
Concretely, given a $f_q$ 
and audio-visual features $\{f_a^t,f_v^t\}_{t=1}^{T}$, the temporal grounding module will learn to aggregate question-aware audio and visual features. The grounded audio feature $\bar{f}_a$ and visual feature $\bar{f}_v$ can be computed as:
\vspace{-0.3em}
\begin{align}
    \bar{f}_a = \sum_{t=1}^{\intercal}w_t^af_a^t={\sigma}(\frac{f_q\textbf{\textit{f}}_a^{\intercal}}{\sqrt{d}})\textbf{\textit{f}}_a\enspace,\\
    \bar{f}_v = \sum_{t=1}^{\intercal}w_t^vf_v^t={\sigma}(\frac{f_q\textbf{\textit{f}}_v^{\intercal}}{\sqrt{d}})\textbf{\textit{f}}_v\enspace,
\end{align}
where $\textbf{\textit{f}}_a = [f_a^1;...;f_a^{T}]$ and $\textbf{\textit{f}}_v= [f_v^1;...;f_v^{T}]$; $d$ is a scaling factor with the same size as the feature dimension. Obviously, the model will assign large weights to audio and visual segments, which are more relevant to the asked question. 
Hence, the question grounded audio/visual contextual embeddings are more capable of predicting correct answers.

\vspace{-0.2em}
\subsection{Multimodal Fusion and Answer Prediciton}
\vspace{-0.4em}
\label{sec:MF}
Different modalities can contribute to correctly answer questions. To combine the features: $\bar{f}_a$, $\bar{f}_v$, and $f_q$, we introduce a simple multimodal fusion network. 
It firstly concatenates audio and visual features and then uses a linear layer with a tanh activation to generate an audio-visual embedding $f_{av}$. Finally, we integrate audio-visual and question features with employing an element-wise multiplication operation. Concretely, we can formulate the fusion function as: 
$e = f_{av} \circ f_q$, where $f_{av} = \text{\textbf{FC}}(\text{Tanh}(\text{Concat}[\bar{f}_a, \bar{f}_v]))$. 

To achieve audio-visual video question answering, we predict the answer for a given question from the joint multimodal embedding $e$. It can be formulated as an open-ended task, which aims to choose one correct word as the answer from a pre-defined answer vocabulary. We utilize a linear layer and softmax function to output a probabilities $p \in \mathcal{R}^{C}$ for candidate answers. With the predicted probability vector and the corresponding ground-truth label $y$, we can optimize our network using a cross-entropy loss: $\mathcal{L}_{qa} = -\sum_{c=1}^{C} y_clog(p_c)$. During testing, we can select the predicted answer by $\hat{c} = \arg \text{max}_c(p)$.



\section{Experiments}
\label{sec:Experiments}
\vspace{-0.2em}
\subsection{Experiments Setting}
\vspace{-0.4em}

\noindent \textbf{Implementation Details.}
 The sampling rates of sounds and video frames are 16 $kHz$ and 1 $fps$, respectively. For each video, we divide it into non-overlapping segments of the same length with 1 frame and generate a 512-D feature vector for each visual segment. For each 1$s$-long audio segment, we use a linear layer to process the extracted 128-D VGGish feature into a 512-D feature vector. The dimension of the word embedding is set to 512. In experiments, due to the limitation of computing resources, we sampled the videos by taking 1$s$ every 6$s$. Batch size and number of epochs are 64 and 30, respectively. The initial learning rate is 1$e$-4 and will drop by multiplying 0.1 every 10 epochs. Our networks is trained with the Adam optimizer. 

\begin{table*}[h]
\begin{center}

\scalebox{0.64}{
\begin{tabular}{c|c|ccc|ccc|cccccc|c}
\hline
\multirow{2}{*}{Task}     & \multirow{2}{*}{Method} & \multicolumn{3}{c|}{Audio Question} & \multicolumn{3}{c|}{Visual Question} & \multicolumn{6}{c|}{Audio-Visual Question}                         & All   \\
                          &                         & Counting   & Comparative   & Avg.   & Counting   & Location   & Avg.    & Existential & Location & Counting & Comparative & Temporal & Avg.  & Avg.  \\ \hline
\multirow{2}{*}{AudioQA}  & FCNLSTM~\cite{fayek2020temporal}                 & 70.45      & 66.22         & 68.88  & 63.89      & 46.74         & 55.21   & \underline{82.01}       & 46.28    & 59.34    & 62.15       & 47.33    & 60.06 & 60.34 \\
                          & CONVLSTM~\cite{fayek2020temporal}                & 74.07      & \textbf{68.89}         & \underline{72.15}  & 67.47      & 54.56         & 60.94   & \textbf{82.91}       & 50.81    & 63.03    & 60.27       & 51.58    & 62.24 & 63.65 \\ \hline
\multirow{4}{*}{VisualQA} & GRU~\cite{antol2015vqa}                     & 72.21      & 66.89         & 70.24  & 67.72      & 70.11         & 68.93   & 81.71       & \underline{59.44}    & 62.64    & 61.88       & 60.07    & 65.18 & 67.07 \\
                          & BiLSTM Attn~\cite{zhou2016attention}             & 70.35      & 47.92         & 62.05  & 64.64      & 64.33         & 64.48   & 78.39       & 45.85    & 56.91    & 53.09       & 49.76    & 57.10 & 59.92 \\
                          & HCAttn~\cite{lu2016hierarchical}                  & 70.25      & 54.91         & 64.57  & 64.05      & 66.37         & 65.22   & 79.10       & 49.51    & 59.97    & 55.25       & 56.43    & 60.19 & 62.30 \\
                          & MCAN~\cite{yu2019deep}                    & \underline{77.50}      & 55.24         & 69.25  & 71.56      & 70.93         & 71.24   & 80.40       & 54.48    & \underline{64.91}    & 57.22       & 47.57    & 61.58 & 65.49 \\ \hline
\multirow{3}{*}{VideoQA}  & PSAC~\cite{li2019beyond}                    & 75.64      & 66.06         & 72.09  & 68.64      & 69.79         & 69.22   & 77.59       & 55.02    & 63.42    & 61.17       & 59.47    & 63.52 & 66.54 \\
                          & HME~\cite{fan2019heterogeneous}                     & 74.76      & 63.56         & 70.61  & 67.97      & 69.46         & 68.76   & 80.30       & 53.18    & 63.19    & 62..69      & 59.83    & 64.05 & 66.45 \\
                          & HCRN~\cite{le2020hierarchical}                    & 68.59      & 50.92         & 62.05  & 64.39      & 61.81         & 63.08   & 54.47       & 41.53    & 53.38    & 52.11       & 47.69    & 50.26 & 55.73 \\ \hline
\multirow{3}{*}{AVQA}     & AVSD~\cite{schwartz2019simple}            &     72.41     &      61.90      &        68.52       &   67.39     &      74.19      &    70.83           &   81.61      &    58.79         &  63.89        &    61.52                 &    61.41      &    65.49   &   67.44    \\
                          & Pano-AVQA~\cite{yun2021pano}               & 74.36      & 64.56         & 70.73  & \underline{69.39}      & \underline{75.65}         & \underline{72.56}   & 81.21       &  59.33   & \underline{64.91}    & \underline{64.22}       & \underline{63.23}    & \underline{66.64} & \underline{68.93} \\ \cline{2-15} 
                          & Our method                & \textbf{78.18}      & \underline{67.05}         & \textbf{74.06}  & \textbf{71.56}      & \textbf{76.38}         & \textbf{74.00}   & 81.81       &  \textbf{64.51}    & \textbf{70.80}   & \textbf{66.01}       & \textbf{63.23}    & \textbf{69.54} & \textbf{71.52} \\ \hline
\end{tabular}
}
\vspace{-0.5em}
\caption{AVQA results of different methods on the test set of MUSIC-AVQA. The top-2 results are highlighted.}
\vspace{-2.2em}
\label{cmp}
\end{center}
\end{table*}


 \noindent \textbf{Training Strategy.} We use a two-stage training strategy, training the spatial grounding module first with $\mathcal{L}_{s}$. Later, based on stage one, using $\mathcal{L}=\mathcal{L}_{qa}+\lambda \cdot \mathcal{L}_{s}$ to train for AVQA task, where $\lambda$ is 0.5 in our experiment.
 
 
\begin{table}
\begin{center}

\scalebox{0.78}{
\begin{threeparttable}
\begin{tabular}{c|cccc}
\hline
Method         & A Question & V Question & A-V Question & All   \\ \hline
Q              & 65.19      & 44.42      & 55.15        & 54.09 \\
A+Q            & 67.78      & 62.75      & 63.86        & 64.26 \\
V+Q            & 68.76      & 67.28      & 63.23        & 65.28 \\
AV+Q           & 70.67      & 69.72      & 65.84        & 67.72 \\
AV+Q+TG        & 73.01      & 73.18      & 68.02        & 70.27 \\
AV+Q+TG+SG & 74.06      & 74.00      & 69.54        & 71.52 \\ \hline
\end{tabular}
\begin{tablenotes}  
        \item[*] TG: Temporal Grounding; SG: Spatial Grounding.
      \end{tablenotes}        
\end{threeparttable}
}
\vspace{-0.7em}
\caption{Ablation study on input modalities and the proposed modules. We observe that leveraging audio, visual, and question information can boost AVQA task.}
\vspace{-2.8em}
\label{tbl:ablation}
\end{center}
\end{table}

 \noindent \textbf{Baselines.} To validate our method on the released MUSIC-AVQA dataset, we compare it with recent audio QA methods: FCNLSTM~\cite{fayek2020temporal} and CONVLSTM~\cite{fayek2020temporal}, visual QA methods: GRU~\cite{antol2015vqa}, BiLSTM Attn~\cite{zhou2016attention}, HCAttn~\cite{lu2016hierarchical} and MCAN~\cite{yu2019deep}, video QA methods: PSAC~\cite{li2019beyond}, HME~\cite{fan2019heterogeneous} and HCRN~\cite{le2020hierarchical}, AVQA method: AVSD~\cite{schwartz2019simple} and Pano-AVQA~\cite{yun2021pano}. To investigate different modalities and modules, we compare several sub-models, as shown in Tab.~\ref{tbl:ablation}.

\noindent \textbf{Evaluation.} We use answer prediction accuracy as the metric and evaluate model performance on answering different types of questions. The answer vocabulary consists of 42 possible answers (22 objects, 12 counting choices, 6 location types, and yes/no) to different types of questions in the dataset. For training, we use one single model to handle all questions without training separated models for each type. So the accuracy with random choice is 1/42$\approx$2.4\%. 
Additionally, all models are trained on our AVQA dataset using the same features for a fair comparison.

\subsection{Results and analysis}
\vspace{-0.4em}
To study different input modalities and validate the effectiveness of the proposed model, we conduct extensive ablations of our model (see Tab.~\ref{tbl:ablation}) and compare to recent QA approaches (see Tab.~\ref{cmp}).

\noindent \textbf{Question-only baseline.} Table~\ref{tbl:ablation} shows the results of the ablation study. The model Q, which only use questions as inputs, achieves accuracy of 54.90, since some type of questions can be answered fully based on common sense. This a common phenomenon that exists in the QA dataset~\cite{antol2015vqa,zhang2016yin,yun2021pano}. For example, on Pano-AVQA dataset~\cite{yun2021pano}, the model Q even outperforms AVSD~\cite{schwartz2019simple} method. However, the model Q is limited in handling complicate QA tasks (\eg, \textit{Location} and \textit{Temporal}). After modeling the spatial and temporal association across modalities, the model performance gains a considerable improvement.




\begin{figure*}[t]
     \centering
     \includegraphics[width=0.97\textwidth]{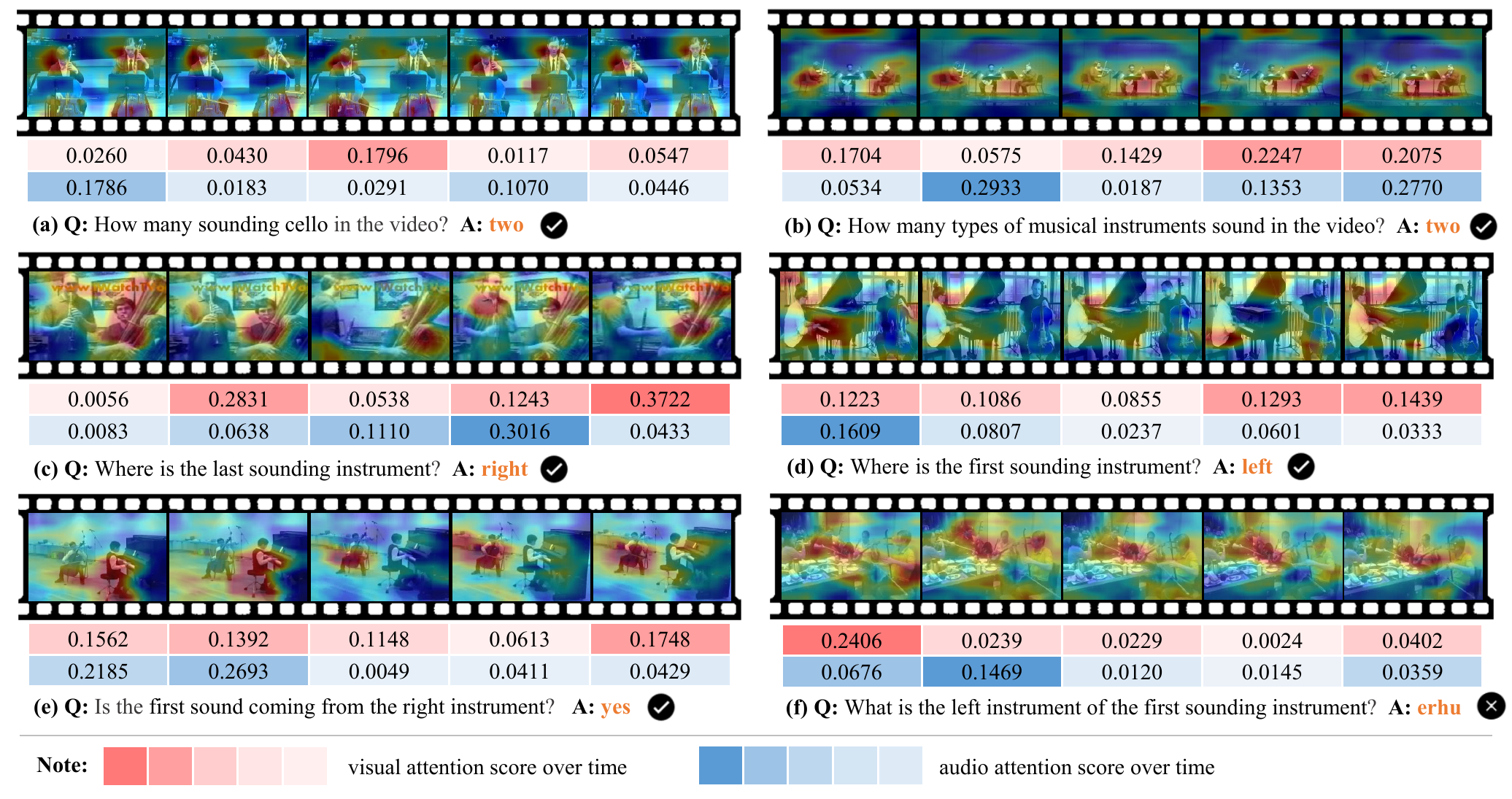}
    \vspace{-1.1em}
     \caption{\textbf{Visualized spatio-temporal grounding results.} Based on the grounding results of our method, the sounding area and key timestamps are accordingly highlighted in spatial and temporal perspectives (a-e), respectively, which indicates that our method can model the spatio-temporal association over different modalities well, facilitating the scene understanding and reasoning. Besides, the subfigure (f) shows one failure case predicted by our method, where the complex scenario with multiple sounding and silent objects makes it difficult to correlate individual objects with mixed sound, leading to a wrong answer for the given question.} 
     \label{fig:spatial}
      \vspace{-1.7em}
\end{figure*}

\noindent \textbf{Multisensory perception boosts QA.} As shown in Tab.~\ref{tbl:ablation}, introducing A or V both facilitates the model performance. Also, the model V+Q adding visual features is overall better than the Q and the A+Q, which indicates that the visual modality is a strong signal for QA. It is not surprising to see that the V+Q is better than A+Q for visual question answering, but we also observe that V+Q outperforms A+Q for audio question answering. 
It is intuitive that recognizing sounds from complicated sound mixtures are very challenging, especially when two sounds are in the same category, while it is easy for visual modality since different sources are visually isolated. As shown in Fig.~\ref{fig:spatial}(a) shows, there are two sounding cellos in the video, which can be seen in visual effortlessly, while the sound of two trumpets is hard to recognized. What's more, obviously, when combining audio and visual modalities, the AV+Q model performance is much better than the A+Q and V+Q models, indicating that multisensory perception helps to boost QA performance.

\noindent \textbf{Spatio-temporal grounding analysis.} With the spatio-temporal grounding module, our audio-visual model achieves the overall best performance among the compared methods. In Fig.~\ref{fig:spatial}, we provide several visualized spatial grounding results. The heatmap indicates the location of sounding source. Through the spatial grounding results, the sounding objects are visually captured, which can facilitate the spatial reasoning. For example, in the case of Fig.~\ref{fig:spatial}(c), the spatial grounding module offers the information that the sounding object in each timestamp. Also, the temporal grounding module aggregate the information of all timestamps based on the question. According to the keyword: \textit{last}, the model can infer that at the last of the video, the instrument located on the right is playing. Combined with temporal grounding module, the model can capture the sounding objects in each timestamp and have a comprehensive understanding of the whole video.

\noindent \textbf{Comparison to recent QA methods.} 
Table~\ref{cmp} shows results of recent QA methods on our MUSIC-AVQA dataset. 
The results firstly demonstrate that all AVQA methods outperform A-, V- and VideoQA methods, which indicates that AVQA task can be boosted through multisensory perception. 
Secondly, our method achieves considerable improvement on most audio and visual questions. 
For the audio-visual question that desires spatial and temporal reasoning, our method is clearly superior over other methods on most question types, especially on answering the \emph{Counting} and \emph{Location} questions.
Although the Pano-AVQA~\cite{yun2021pano} attempted to model audio-visual scenes, our methods explicitly constructs the association between audio and visual modalities and temporally aggregate both features, solving the spatio-temporal reasoning problem more effectively. 
Moreover, the results confirm the potential of our dataset as a testbed for audio-visual scene understanding.


\vspace{-0.5em}
\label{sec:discussion}
\section{Discussion}
\vspace{-0.5em}


In this work, we investigate the audio-visual question answering problem, which aims to answer questions regarding videos by fully exploiting multisensory content. To facilitate this task, we build a large-scale MUSIC-AVQA dataset, which consists of 45,867 question-answer pairs spanning over audio-visual modalities and different question types. We also propose a spatio-temporal grounding model to explore the fine-grained scene understanding and reasoning. Our results show that all of different modalities can contribute to addressing the AVQA task and our model outperforms recent QA approaches, especially when equipped with our proposed modules. We believe that our dataset can be a useful testbed for evaluating fine-grained audio-visual scene understanding and spatio-temporal reasoning, and has a potential to inspire more people to explore the field.

\noindent  \textbf{Limitation.} 
Although we have achieved considerable improvement, the AVQA task still has a wide scope for exploration. 
Firstly, the scene of the current dataset is more limited to the musical scenario, while audio-visual interaction exists in more daily situations. We will explore audio-visual reasoning tasks in more general scenarios in the subsequent study.
Our model simply decomposes the complex scenarios into concrete audio-visual association. However, some visual objects or sound sources, which are not relevant to the questions, are involved in the encoded unimodal embeddings, might introducing learning noises and make solving QA tasks challenging, as the shown failure example in Fig.~\ref{fig:spatial}(f). 
To alleviate the problem, we can parse each video into individual objects and isolated sounds and then adaptively leverage question-related audio and visual elements for more accurate question answering. 
Further, to facilitate temporal reasoning, we proposed to highlight the key timestamps that are close to the question. 
However, such module lacks explicit temporal modeling between audio and visual modality. 
More advanced model that could bridge the temporal association across modalities is expected to boost performance further. 
Though the scenarios are somewhat limited, we think this is the first step of audio-visual reasoning and we believe this paper will be a good start in this field.

\noindent
\textbf{Broader impacts.} The released MUSIC-AVQA dataset is curated, which perhaps owns potential correlation between instrument and geographical area. This issue warrants further research and consideration.

\noindent
\textbf{Acknowledgement}
G. Li, Y. Wei, J-R. Wen and D. Hu were supported by Intelligent Social Governance Platform, Major Innovation \& Planning Interdisciplinary Platform for the ``Double-First Class'' Initiative, Renmin University of China. They were also supported by Beijing Outstanding Young Scientist Program (NO.BJJWZYJH012019100020098), the Research Funds of Renmin University of China (NO.21XNLG17), the National Natural Science Foundation of China (NO.62106272), the 2021 Tencent AI Lab Rhino-Bird Focused Research Program (No.JR202141), the Young Elite Scientists Sponsorship Program by CAST, the Large-Scale Pre-Training Program of Beijing Academy of Artificial Intelligence (BAAI) and the Public Computing Cloud, Renmin University of China. Y. Tian and C. Xu were supported by the National Science Foundation (NSF) under Grant 1741472. The article solely reflects the opinions and conclusions of its authors but not the funding agents.

{\small
\bibliographystyle{ieee_fullname}
\bibliography{egbib}
}

\appendix
\clearpage




\section{Supplementary Video}
\vspace{-0.4em}
In our demo video, we will provide video examples with sounds in our MUSIC-AVQA dataset and audio-visual question answering results. For more details, please check the demo.

\section{Videos Collection}
\vspace{-0.4em}
In this section, We introduce the details of MUSIC-AVQA dataset construction. 
According to \textit{Wikipedia}, 22 kinds of instruments shown in Tab.~\ref{tab:musical} are divided into 4 categories: \textit{String}, \textit{Wind}, \textit{Percussion} and\textit{ Keyboard}.

\begin{table}[ht]
\begin{center}
\vspace{-0.4em}
\caption{Musical Instrument Classification}
\vspace{-0.8em}
\label{tab:musical}
\begin{tabular}{c|c|c|c}
\midrule
String & Wind & Percussion & Keyboard\\
\midrule
violin&tuba&drum&accordion\\
cello&trumpet&xylophone&piano\\
guitar&suona&congas&\\
ukulele&bassoon&&\\
erhu&clarinet&&\\
guzheng&bagpipe&&\\
pipa&flute&&\\
bass&saxophone&&\\
banjo&&&\\
\bottomrule
\end{tabular}
\end{center}
\vspace{-1.5em}
\end{table}

\subsection{Real Videos}
\vspace{-0.4em}
In the MUSIC-AVQA dataset, three kinds of musical performance are covered to ensure the diversity, complexity and dynamic of audio-visual scenes: solo, ensemble of the same instrument (ESIT) and ensemble of different instruments (EDIT).
The rule of EDIT is that each instrument is required to combine with one or more instruments in different categories.
Specifically, we use permutation and combination methods for 22 instruments to ensure that all instrument combinations can be covered in the video as much as possible.
For the duet case in EDIT, we consider all the combinations of 2 different categories of 22 instruments, which accordingly becomes a total of $C_{22}^2$ combinations.
We search for related videos on YouTube according to these combinations styles.
Meanwhile, for other ensemble forms in EDIT, such as trio, quartet, etc., we consider more than 2 different instrument combinations and retrieve related videos on YouTube.

\begin{figure*}[ht]
     \centering\textbf{}
     \includegraphics[width=0.95\textwidth]{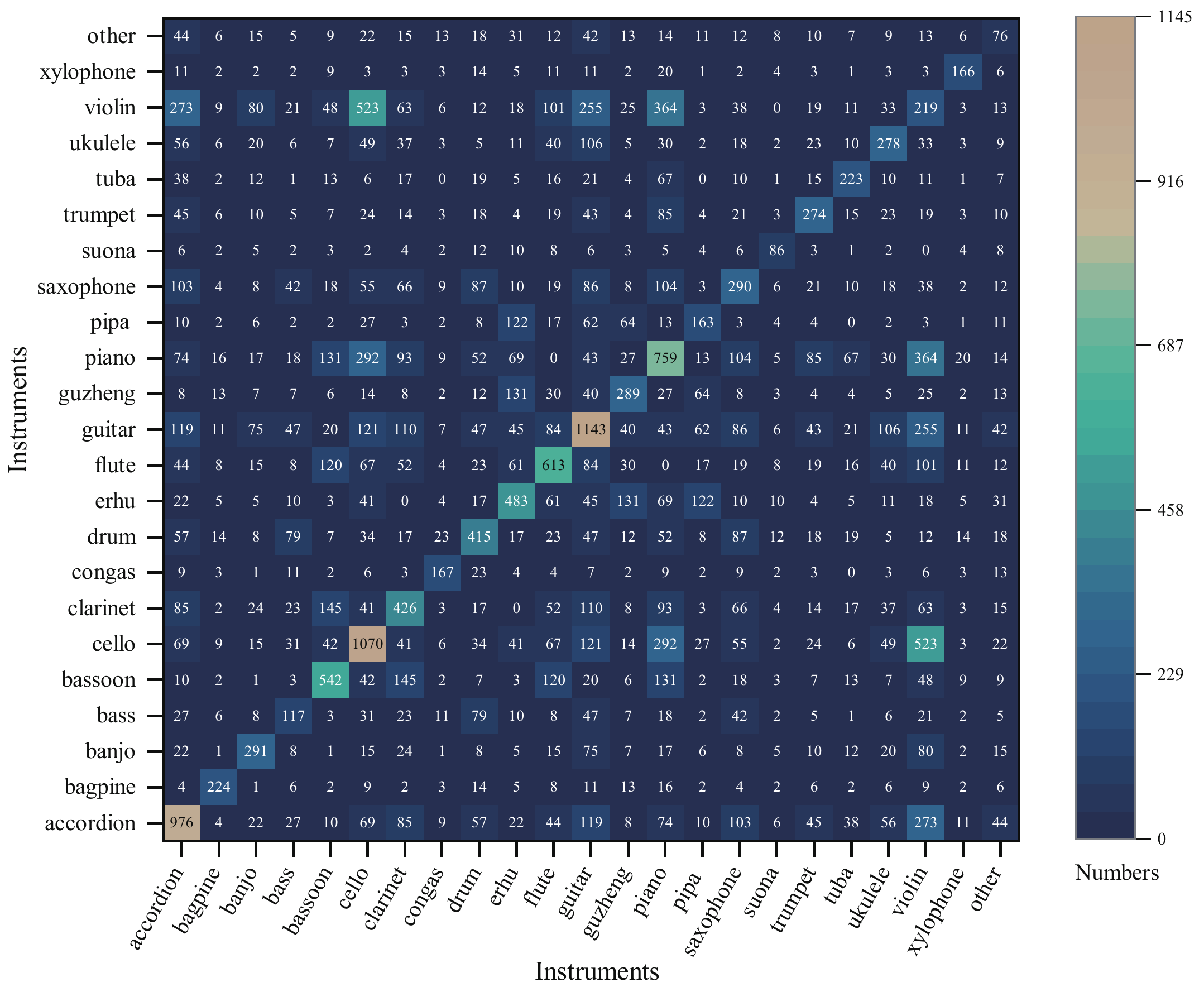}
     \vspace{-1em}
     \caption{Number of combinations of different types of instruments, where the lighter the color, the more the number.
     And instruments outside the 22 instrument categories are denoted by \textit{other}.
     The confusion matrix shows that the combination of different instruments is diversified.}
     \label{fig:matrix}
     \vspace{-1.2em}
\end{figure*}

In Fig.~\ref{fig:matrix}, we show the number of the combination of every two different instruments in the real video, counted from not only the duet video, but also the trio, quartet etc.
The categories of musical instruments appearing in some videos are not in the 22 musical instruments and which are represented by \textit{other}. As shown in Fig.~\ref{fig:matrix}, some instruments tend to combine with some other instruments due to their coordination in music, such as \textit{cello} and \textit{violin} etc. 
Even though, we still do our best to find almost all kinds of combination of different instruments. 
These statistical results illustrate the diversity of the collected videos.

\subsection{Synthetic Videos} 
\vspace{-0.5em}
\begin{figure*}[htb]
     \centering
     \includegraphics[width=0.90\textwidth]{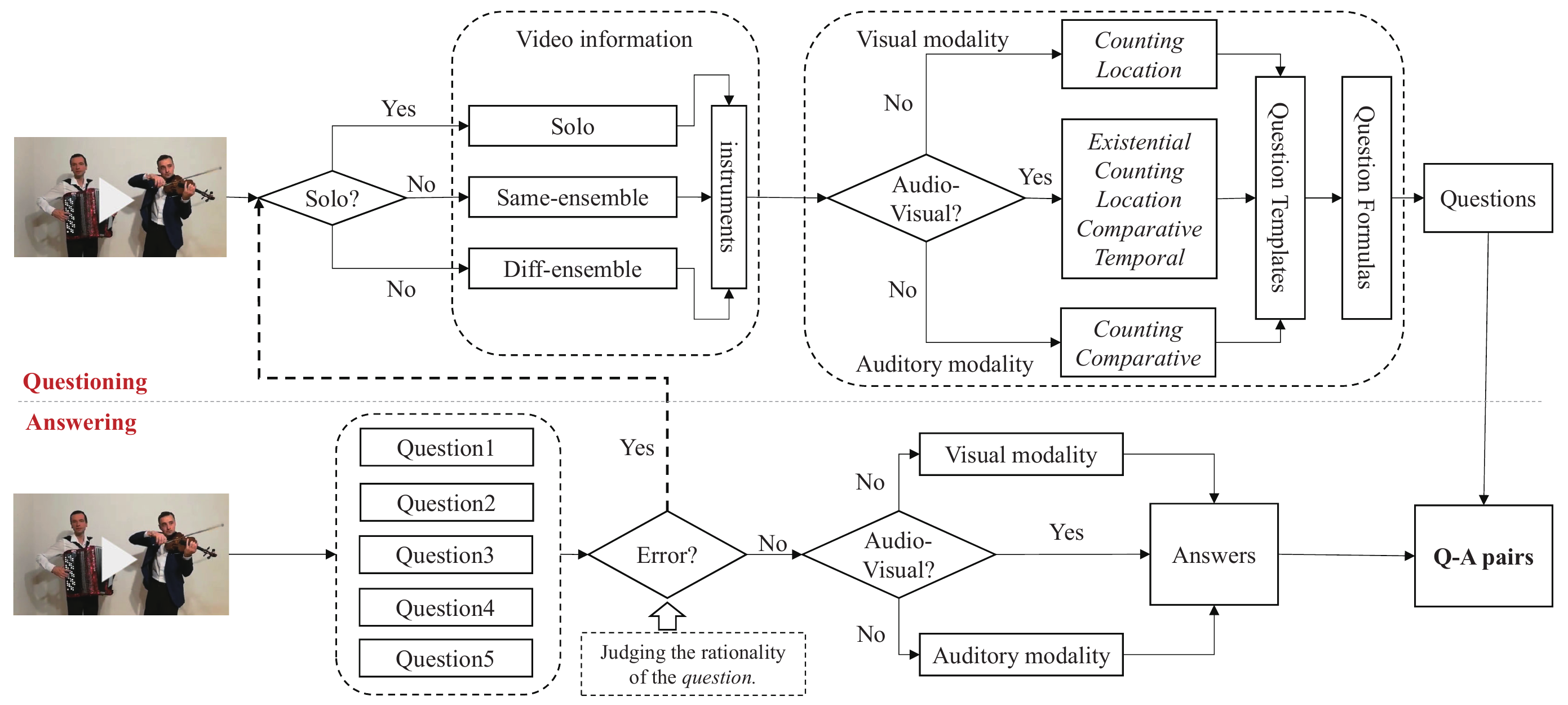}
     \vspace{-4mm}
     \caption{Labeling system contains \textit{questioning} and \textit{answering}. 
     In the \textit{questioning} section, the annotator is required to select the performance type of the video and the included instruments, and then \textit{scene type}s, \textit{question types}, and \textit{question templates}, and finally one \textit{question} is automatically generated based on the previous selection.
     In the \textit{answering} part, the annotator to judge whether the \textit{question} is reasonable, and if it is unreasonable, the \textit{question} will be labeled again.
     Then, the annotator answering the \textit{question} according to video content, and finally one QA pair is produced.}
     \label{fig:flow}
    \vspace{-3mm}
\end{figure*}

To further facilitate study on understanding and reasoning over complex multimodal scenes, we synthesize more challenging videos in which multiple visual objects and sounds are appeared with different associations. 

For videos synthesized using solo scenes, we retrieve about another 1,500 videos from YouTube, w.r.t. above 22 instrument categories, and they are not included in the collected solo videos in Sec. \textcolor[rgb]{1,0,0}{2.1} above.

Additionally, the number of solo videos for each instrument is between 50-80, and all the 1500 videos are randomly cut into 1 minute long.
For simplicity, the cutted video is denoted as $D$.
Then, we randomly select 750 videos from \textit{D} and separate the sound track from them. 
The separated video (silent) and audio are represented by $D_V$ and $D_A$, respectively.
After that, we divide $D$ into two types: $M$ and $N$, where $M$ contains $D_V$ and $D_A$, and the rest videos in $D$ except $M$ is represented by $N$.
Finally, we synthesize videos in the following three ways.

\textbf{1) Audio overlay}.
We randomly select 500 audios and videos from $D_A$ and $N$, respectively.
Then we randomly select one audio and overlay it to one video, which generate one video contain single instrument in vision but with two instrument sounds.

\textbf{2) Video stitching}.
We randomly select two different real videos then spatially stitch them into one video.
Specially, we select 500 videos from $D_A$ and $N$, respectively.
Then these two different types of videos are randomly stitched horizontally into one video, so that one video will contain the left and right instrument performance, but only one of them has sound.

\textbf{3) Audio and video random matching}.
We replace the original sound of real videos with the sound track from another randomly selected video. In details, 500 samples are randomly selected from $D_A$ and $D_V$, respectively. Then the audio in $D_A$ is randomly superimposed on a video in $D_V$, hence the instrument and sound in the video do not match.

In addition, we also employ the above synthesizing operation on the ensemble videos,
where about 1,000 videos are collected in the same way as ESIT and EDIT in Sec. \textcolor[rgb]{1,0,0}{2.1}, but the collected videos are not in the videos in Sec. \textcolor[rgb]{1,0,0}{2.1}.
Finally, a total of about 1,866 synthetic videos are obtained, which constitutes the whole musical performance video set with the real-world ones.

\section{QA pair Collection}
\subsection{Questions Design}
\vspace{-0.4em}
In different modality scenarios, 33 question templates covering 9 question types are proposed.
Tab.~\ref{tab:questiontype} shows 9 question types in different scenarios, and the specific 33 question templates are given in Sec. \textcolor[rgb]{1,0,0}{7}.

\begin{table}[htp]
\begin{center}
\vspace{-0.4em}
\caption{Three scenarios and their corresponding question types.}
\vspace{-0.8em}
\label{tab:questiontype}
\begin{tabular}{c|c|c}
\midrule
Audio-Visual & Visual & Audio\\
\midrule
Existential &  & \\
Counting & Counting & Counting\\
Location & Location & Comparative\\
Comparative &   &  \\
Temporal &   &  \\
\bottomrule
\end{tabular}
\end{center}
\vspace{-1.5em}
\end{table}

\subsection{QA pairs Collection}
\vspace{-0.4em}
We design an audio-visual question answering labeling system to collect questions, and all QA pairs are collected with this system.
The flow chart of the labeling system is shown in Fig.~\ref{fig:flow}.
First, questions are required to raise w.r.t three different modality scenarios, namely \textit{Audio-Visual}, \textit{Visual} and \textit{Audio}, to explore the different modal contents. Then, for each modality scenario, different question types are designed to meet the requirements of scene understanding and reasoning, such as \textit{existential, counting, location}, etc. At last, for each question type, we design multiple question templates that consist of fixed sentence pattern and formulas.

\subsection{QA pairs samples}
\begin{figure*}[t]
     \centering
     \includegraphics[width=0.99\textwidth]{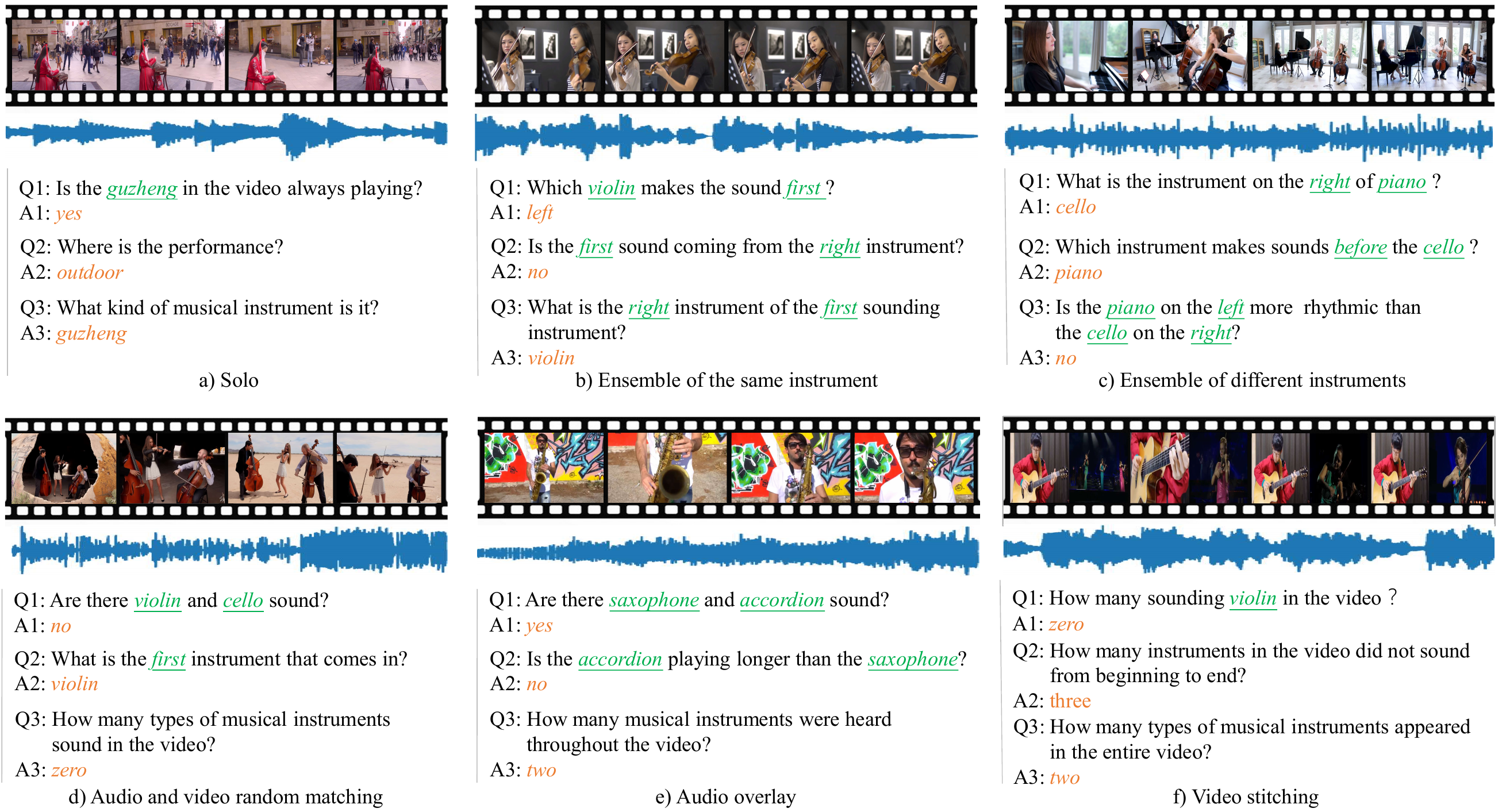}
     \caption{Different audio-visual scene types and their annotated QA pairs in the AVQA dataset. In the first row, a), b), and c) represent real musical performance videos, namely \textit{solo}, \textit{ensemble of the same instrument}, and \textit{ensemble of different instruments}. In the second row, d), e), and f) represent the synthetic video, which are \textit{audio and video random matching}, \textit{audio overlay}, and \textit{video stitching}, respectively. }
     \label{fig:dataset_demo}
    \vspace{-4mm}
\end{figure*}

\vspace{-0.4em}
The large-scale spatial-temporal audio-visual dataset that focuses on question-answering task, as shown in Fig.~\ref{fig:dataset_demo}

\section{Auxiliary experiments}

\subsection{Temporal modeling with shuffled segments.}
\vspace{-0.4em}
To better evaluate the Temporal Grounding (TG) module and answer the question, we exclude the Spatial Grounding module and shuffle each input video in the time dimension of AV+Q+TG model. 
Without shuffling, the performance on the temporal questions is 65.17 while the performance drops to 63.71 after shuffling. 
Since the TG module does not explicitly encode the temporal order information of videos, shuffling the video segments does not affect the performance a lot. 
But the model with correct temporal information still achieves better on Temporal questions. 
One possible reason is that temporal-related words in questions, such as first and last, can implicitly help the model group to the corresponding temporal location. 
To further improve temporal question answering and strengthen temporal reasoning capability of our framework, it would be interesting to explore explicitly utilizing the temporal order information from the two modalities in the future.

\subsection{Modeling with motion information}
To further utilize the temporal information of the video, we use R(2+1)D network to extract motion features, which are fused to visual features. Our method with motion information achieves 71.75 on the released MUSIC-AVQA dataset, which is better than our method (71.53). According to the results, the model performance is boosted when combining motion information.

\subsection{Experiments on existing video QA dataset}

To explore whether the existing video QA dataset is suitable for AVQA task, we conduct experiments on the TVQA dataset~\cite{lei2018tvqa}, a large-scale video QA dataset based on 6 popular TV shows. Since the original TVQA framework does not take audio information as input, we add an audio encoder, a pre-trained VGGish\cite{gemmeke2017audio} model, to extract audio features. Also, to be fair, we only take the ImageNet features as the video input, and the temporal dimension of question/answer features are squeezed by average operation. Different inputs are taken to comparison.

As the results shown in Tab.~\ref{tbl:tvqa}, both Q+V and Q+V+A methods are not superior to Q-only method based on common sense, which is consistent with the results reported in TVQA~\cite{lei2018tvqa}.
In addition, our method outperforms TVQA method in both visual-only and audio-visual inputs. But the introduced audio modality harms the performance of both methods. We consider the reason is that the sound in TVQA dataset is mainly human speech~\cite{lei2018tvqa}, and it is hard to modelling the interaction across
\begin{wraptable}{rb}{4.0cm}
\vspace{-1mm}
\begin{center}
\vspace{-3mm}
\caption{\textbf{Experiments on TVQA dataset.} Q: Question. V: Video. A: Audio. *: TVQA method. \dag: Our method.}
\vspace{-3mm}
\label{tbl:tvqa}
\begin{tabular}{c|c}
\hline
\textbf{Method} & \textbf{Accuracy} \\ 
\hline
Q-only*   & 43.50 \\
Q+V*           & 41.70 \\
Q+V+A*       &  41.45  \\ \hline
Q+V\dag   &    42.01     \\ 
Q+V+A\dag  &    41.95      \\ 
\hline
\end{tabular}
\end{center}
\vspace{-3mm}
\end{wraptable}
 both modalities. 
This phenomenon indicates that TVQA dataset is not quite suitable for the AVQA task which needs to explore the interactions between audio and visual components.
In such a situation, our method still shows better robustness with less performance drop.

\section{Examples}
\begin{figure}[!h]
     \centering
     \includegraphics[width=0.46\textwidth]{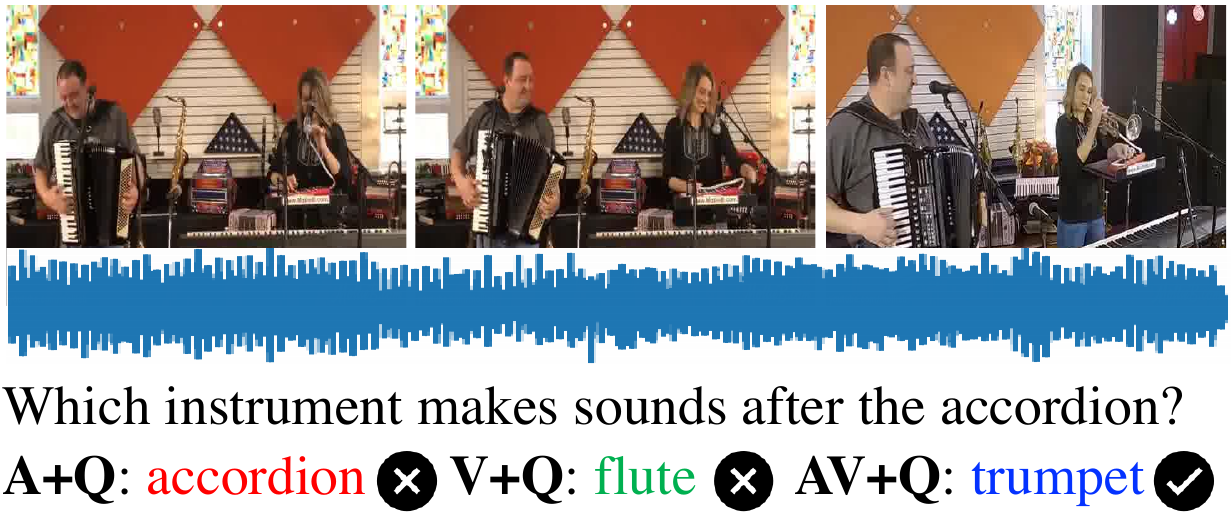}
     \caption{Our audio-visual model predicts the correct answer but the individual audio and visual models fail. To answer this question, the model needs to perform multimodal scene understanding and temporal reasoning over the video.}
     \label{fig:AvvsSingle}

\end{figure}

\begin{figure*}[!ht]
     \centering
     \includegraphics[width=0.99\textwidth]{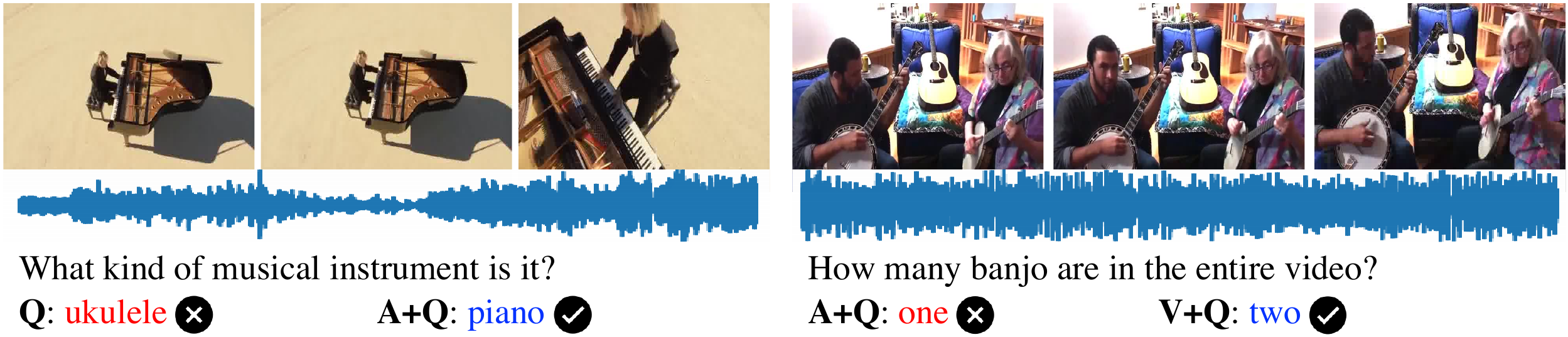}
     \caption{Ablation on input modalities. Left: leveraging the audio modality, the model A+Q can answer the correct instrument \textit{piano} in the video. Right: with the help of the visual modality, V+Q recognizes two banjos in the video. However, the A+Q gives an wrong answer since it is more difficult to distinguish the number of sound sources in the same category for the audio.}
     \label{fig:AvsQ}
\end{figure*}

\begin{figure*}[!hb]
     \centering
     \includegraphics[width=0.99\textwidth]{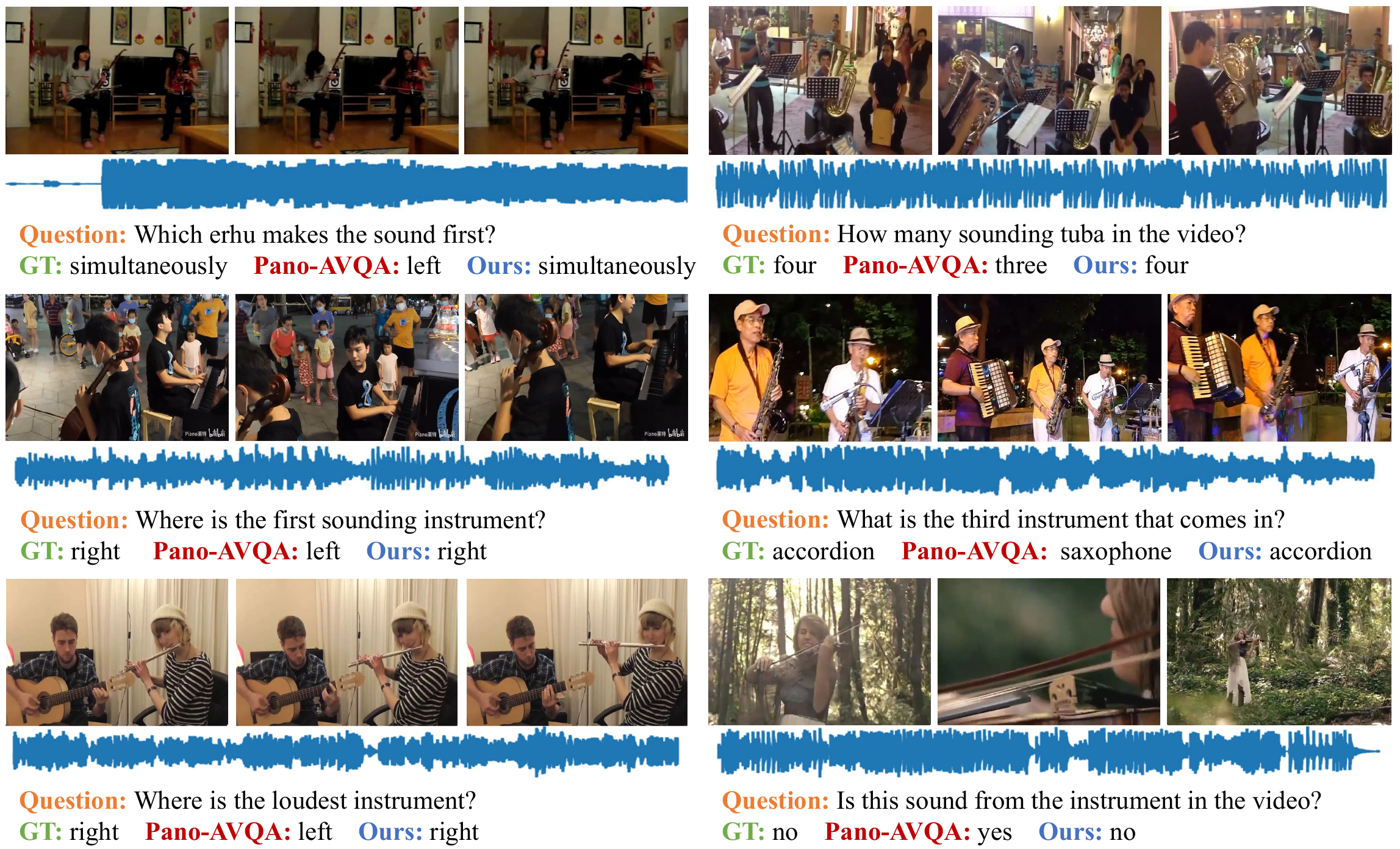}
     \caption{Audio-visual question answering results. Our model can predict correct answers to the questions and is better than the recent AVQA method: Pano-AVQA~\cite{yun2021pano}.}
     \label{fig:res}
\end{figure*}

To further study different input modalities and validate the effectiveness of the proposed model and compare to recent QA methods, we visualize some QA examples and have following findings:

First, audio improves question answering. The left example in Fig.~\ref{fig:AvsQ} shows that the additional audio modality helps our model to answer the question. With the assistance of audio, the model can distinguish which instrument is playing. Second, visual modality is crucial. The visual modality is a strong signal for QA. One example is illustrated in the right of Fig.~\ref{fig:AvsQ}. In this case, recognizing sounds from complicated sound mixtures are very challenging, especially when two sounds are in the same category, while different sources are naturally isolated in the visual modality. The interesting results can support that auditory scene understanding can also benefit from visual perception. Third, multisensory perception boosts QA. An example is shown in Fig.~\ref{fig:AvvsSingle}. With recognizing sounding scenes and performing temporal reasoning, our audio-visual model can identify the sounding instrument trumpet after the accordion. From the results, we can learn that the two different modalities contain complementary information and multisensory perception is helpful for the fine-grained scene understanding task. Last but not least, to validate effectiveness of the proposed method, we compare it to a recent AVQA method: Pano-AVQA~\cite{yun2021pano}. Several samples are provided in Fig.~\ref{fig:res}. We can find that our method, which explicitly constructs the association between audio and visual modalities and temporally aggregates audio and visual features, can predict correct answers to the questions and obtains superior performance.

\section{Personal data/Human subjects}

Videos in MUSIC-AVQA are public on YouTube, and annotated via crowdsourcing. We have explained how the data would be used to crowdworkers. Our dataset does not contain personally identifiable information or offensive content.

\section{Question Templates}
The 33 question templates in the AVQA dataset are shown in Table~\ref{question_tupes}.
\begin{table*}[ht]
\begin{center}
\caption{The 33 question templates.}
\label{question_tupes}
\scalebox{0.9}{
\begin{tabular}{ccl}
\hline
\multicolumn{1}{c|}{\textbf{Modalities}}                     & \multicolumn{1}{c|}{\textbf{Question Types}}               & \multicolumn{1}{c}{\textbf{Question Templates}}                                                                                                                                                           \\ \hline
\multicolumn{1}{c|}{\multirow{18}{*}{Audio-Visual}} & \multicolumn{1}{c|}{\multirow{3}{*}{Existential}} & Is this sound from the instrument in the video?                                                                                                                                                  \\
\multicolumn{1}{c|}{}                               & \multicolumn{1}{c|}{}                             & Is the $<$\textit{Object}$>$ in the video always playing?                                                                                                                       \\
\multicolumn{1}{c|}{}                               & \multicolumn{1}{c|}{}                             & Is there a voiceover?                                                                                                                                                                            \\ \cline{2-3} 
\multicolumn{1}{c|}{}                               & \multicolumn{1}{c|}{\multirow{4}{*}{Counting}}    & How many instruments are sounding in the video?                                                                                                                                                  \\
\multicolumn{1}{c|}{}                               & \multicolumn{1}{c|}{}                             & How many types of musical instruments sound in the video?                                                                                                                                        \\
\multicolumn{1}{c|}{}                               & \multicolumn{1}{c|}{}                             & How many instruments in the video did not sound from beginning to end?                                                                                                                           \\
\multicolumn{1}{c|}{}                               & \multicolumn{1}{c|}{}                             & How many sounding $<$\textit{Object}$>$ in the video?                                                                                                                           \\ \cline{2-3} 
\multicolumn{1}{c|}{}                               & \multicolumn{1}{c|}{\multirow{4}{*}{Location}}    & Where is the $<$\textit{LL}$>$ instrument?                                                                                                                                      \\
\multicolumn{1}{c|}{}                               & \multicolumn{1}{c|}{}                             & Is the $<$\textit{FL}$>$ sound coming from the $<$\textit{LR}$>$ instrument?                                                                                   \\
\multicolumn{1}{c|}{}                               & \multicolumn{1}{c|}{}                             & Which is the musical instrument that sounds at the same time as the $<$\textit{Object}$>$?                                                                                      \\
\multicolumn{1}{c|}{}                               & \multicolumn{1}{c|}{}                             & What is the $<$\textit{LR}$>$ instrument of the $<$\textit{FL}$>$ sounding instrument?                                                                         \\ \cline{2-3} 
\multicolumn{1}{c|}{}                               & \multicolumn{1}{c|}{\multirow{4}{*}{Comparative}} & Is the instrument on the $<$\textit{LR}$>$ more rhythmic than the instrument on the $<$\textit{RL}$>$?                                                         \\
\multicolumn{1}{c|}{}                               & \multicolumn{1}{c|}{}                             & Is the instrument on the $<$\textit{LR}$>$ louder than the instrument on the $<$\textit{RL}$>$?                                                                \\
\multicolumn{1}{c|}{}                               & \multicolumn{1}{c|}{}                             & Is the $<$\textit{Object}$>$ on the $<$\textit{LR}$>$ more rhythmic than the $<$\textit{Object}$>$ on the $<$\textit{RL}$>$? \\
\multicolumn{1}{c|}{}                               & \multicolumn{1}{c|}{}                             & Is the $<$\textit{Object}$>$ on the $<$\textit{LR}$>$ louder than the $<$\textit{Object}$>$ on the $<$\textit{RL}$>$?        \\ \cline{2-3} 
\multicolumn{1}{c|}{}                               & \multicolumn{1}{c|}{\multirow{3}{*}{Temporal}}    & Where is the $<$\textit{FL}$>$ sounding instrument?                                                                                                                             \\
\multicolumn{1}{c|}{}                               & \multicolumn{1}{c|}{}                             & Which $<$\textit{Object}$>$ makes the sound $<$\textit{FL}$>$?                                                                                                 \\
\multicolumn{1}{c|}{}                               & \multicolumn{1}{c|}{}                             & Which instrument makes sounds $<$BA$>$ the $<$\textit{Object}$>$?                                                                                                               \\ \hline
\multicolumn{1}{c|}{\multirow{8}{*}{Visual}}        & \multicolumn{1}{c|}{\multirow{4}{*}{Counting}}    & Is there a $<$\textit{Object}$>$ in the entire video?                                                                                                                           \\
\multicolumn{1}{c|}{}                               & \multicolumn{1}{c|}{}                             & Are there $<$\textit{Object}$>$ and $<$\textit{Object}$>$ instruments in the video?                                                                            \\
\multicolumn{1}{c|}{}                               & \multicolumn{1}{c|}{}                             & How many types of musical instruments appeared in the entire video?                                                                                                                              \\
\multicolumn{1}{c|}{}                               & \multicolumn{1}{c|}{}                             & How many $<$\textit{Object}$>$ are in the entire video?                                                                                                                         \\ \cline{2-3} 
\multicolumn{1}{c|}{}                               & \multicolumn{1}{c|}{\multirow{4}{*}{Location}}    & Where is the performance?                                                                                                                                                                        \\
\multicolumn{1}{c|}{}                               & \multicolumn{1}{c|}{}                             & What is the instrument on the $<$\textit{LR}$>$ of $<$\textit{Object}$>$?                                                                                      \\
\multicolumn{1}{c|}{}                               & \multicolumn{1}{c|}{}                             & What kind of musical instrument is it?                                                                                                                                                           \\
\multicolumn{1}{c|}{}                               & \multicolumn{1}{c|}{}                             & What kind of instrument is the $<$\textit{LRer}$>$ instrument?                                                                                                                  \\ \hline
\multicolumn{1}{c|}{\multirow{6}{*}{Audio}}         & \multicolumn{1}{c|}{\multirow{3}{*}{Counting}}    & Is there a $<$\textit{Object}$>$ sound?                                                                                                                                         \\
\multicolumn{1}{c|}{}                               & \multicolumn{1}{c|}{}                             & How many musical instruments were heard throughout the video?                                                                                                                                    \\
\multicolumn{1}{c|}{}                               & \multicolumn{1}{c|}{}                             & How many types of musical instruments were heard throughout the video?                                                                                                                           \\ \cline{2-3} 
\multicolumn{1}{c|}{}                               & \multicolumn{1}{c|}{\multirow{3}{*}{Comparative}} & Is the $<$\textit{Object1}$>$ more rhythmic than the $<$\textit{Object2}$>$?                                                                                   \\
\multicolumn{1}{c|}{}                               & \multicolumn{1}{c|}{}                             & Is the $<$\textit{Object1}$>$ louder than the $<$\textit{Object2}$>$?                                                                                          \\
\multicolumn{1}{c|}{}                               & \multicolumn{1}{c|}{}                             & Is the $<$\textit{Object1}$>$ playing longer than the $<$\textit{Object2}$>$ ?      \\ \hline                   
\end{tabular}
}
\end{center}
\vspace{-5mm}
\end{table*}


\end{document}